\documentclass[sigconf]{acmart}

\usepackage{xcolor}
\usepackage{booktabs}
\usepackage{stfloats}
\usepackage{balance}

\newcommand{\modified}[1]{\textcolor{black}{#1}}

\AtBeginDocument{%
  \providecommand\BibTeX{{%
    \normalfont B\kern-0.5em{\scshape i\kern-0.25em b}\kern-0.8em\TeX}}}

\copyrightyear{2023} 
\acmYear{2023} 
\setcopyright{acmlicensed}
\acmConference[CHI '23]{Proceedings of the 2023CHI Conference on Human Factors in Computing Systems}{April 23--28, 2023}{Hamburg, Germany}
\acmBooktitle{Proceedings of the 2023 CHI Conference on Human Factors in Computing Systems (CHI '23), April 23--28, 2023, Hamburg, Germany}
\acmPrice{15.00}
\acmDOI{10.1145/3544548.3580893}
\acmISBN{978-1-4503-9421-5/23/04}

\acmSubmissionID{7433}

\begin{document}

\newcommand{\method}{{\textit{sPD}}} 
\newcommand{\methodfull}{{\textit{Situated Participatory Design}}} 

\title[Situated Participatory Design]{Situated Participatory Design: A Method for In Situ Design of Robotic Interaction with Older Adults}

\author{Laura Stegner}
\orcid{0000-0003-4496-0727}
\affiliation{%
  \institution{University of Wisconsin--Madison}
  \city{Madison}
  \state{Wisconsin}
  \country{United States}
  \postcode{53706}
}
\email{stegner@wisc.edu}

\author{Emmanuel Senft}
\orcid{0000-0001-7160-4352}
\affiliation{%
  \institution{Idiap Research Institute}
  \city{Martigny}
  \country{Switzerland}
}
\email{esenft@idiap.ch}

\author{Bilge Mutlu}
\orcid{0000-0002-9456-1495}
\affiliation{%
  \institution{University of Wisconsin--Madison}
  \city{Madison}
  \state{Wisconsin}
  \country{United States}
  \postcode{53706}
}
\email{bilge@cs.wisc.edu}


\begin{abstract}
We present a participatory design method to design human-robot interactions with older adults and its application through a case study of designing an assistive robot for a senior living facility. The method, called \textit{\methodfull{}} (\method{}), was designed considering the challenges of working with older adults and involves three phases that enable designing and testing use scenarios through realistic, iterative interactions with the robot. 
In design sessions with nine residents and three caregivers, we uncovered a number of insights about \method{} that help us understand its benefits and limitations.
For example, we observed how designs evolved through iterative interactions and how early exposure to the robot helped participants consider using the robot in their daily life. With \method{}, we aim to help future researchers to increase and deepen the participation of older adults in designing assistive technologies.

\end{abstract}

\ccsdesc[500]{Human-centered computing~Accessibility design and evaluation methods}
\ccsdesc[500]{Human-centered computing~Participatory design}
\ccsdesc[500]{Computer systems organization~Robotics}
\ccsdesc[500]{Human-centered computing~Field studies}

\keywords{Human-robot interaction, older adults, assistive robots, accessibility, design methods, participatory design, field study}


\begin{teaserfigure}
  \includegraphics[width=\textwidth]{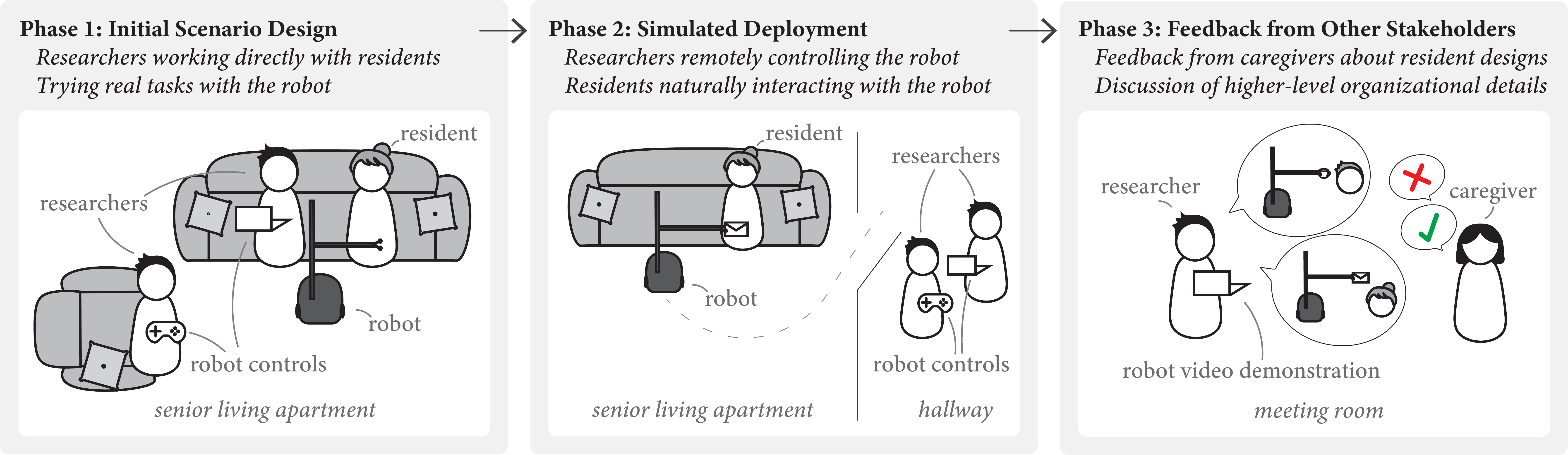}
  \caption{We present \methodfull \ (\method{}), a participatory design (PD) method specially designed to address the challenges of working with older adults to design assistive technologies. \method{} includes three key phases: (1) a \textit{co-design} phase to design an initial scenario; (2) a \textit{simulated deployment} phase to test out the interactions in realistic conditions; and (3) a \textit{follow-up} phase where other stakeholders (\textit{e.g.}, care staff) reflect on resulting designs. We demonstrate the use of \method{} in a case study with the residents and caregivers of a senior living facility and present insights into the benefits of \method{}.}
  \Description{Example of the method through three panels describing the three steps in simple sketches. 
  The first panel is titled ``Phase 1: Initial Scenario Design'', with the subtitle: ``Researchers working directly with residents Trying real tasks with the robot'', and shows two researchers, an older adult sitting together in a resident home and a robot in the middle. The two researchers control the robot which interacts with the older adult. 
  The second panel is titled ``Phase 2: Simulated Deployment'', with the subtitle: ``Researchers remotely controlling the robot
Residents naturally interacting with the robot'', and shows the robot interacting with the  older adult, while the researchers are hidden in the hallway to remotely operate the robot.
The third panel is titled ``Phase 3: Feedback from Other Stakeholders'', with the subtitle: ``Feedback from caregivers about resident designs
Discussion of higher-level organizational details'', and shows an researcher talking to a caregiver in a meeting room and showing them videos of interactions between residents and the robot on a laptop and the caregiver providing feedback on the interactions.
}
  \label{fig:teaser}
\end{teaserfigure}

\maketitle

\section{Introduction}
\label{intro}

Robots are increasingly being used to help older adults live more independently and to overcome a growing shortage of caregivers. Research efforts have focused on addressing a wide range of capabilities and needs, from robots that assist people with limited mobility with bathing \cite{king2010towards} to supporting cognitive and social stimulation \cite{luperto2019towards}. However, despite many technical advances, adoption of robots is still limited \cite{bardaro2021robots}. Recent work has called for increased participation of older adults in the design of assistive technologies to increase their acceptance, usefulness, and adoption \cite{bardaro2021robots,duque2019systematic,czaja2007impact}.

\textit{Participatory Design} (PD) is a method that engages key stakeholders of a product or a service in the design process \cite{lee2017steps}. PD methods enable designers to create personalized systems that help to address the unique needs of specific user groups. Recent research on technology for older adults has successfully used PD to increase the engagement of this population in the design process \cite{duque2019systematic}. This increased engagement can lead to higher acceptance of newer technology by better aligning the design of emerging technologies with the needs and expectations of their users \cite{duque2019systematic}. 

As a general methodology, PD encompasses a wide range of activities, which allows for significant flexibility to craft a specific approach that suits both research questions and participants' needs.
Typical PD activities, \textit{e.g.}, interviews, off-site workshops, and interactions with low-fidelity prototypes, have a low barrier to use and can provide useful insight into the general design of a robot and the specific tasks it can perform. 
Although current PD methods demonstrated promise to address the unique needs of older adults, prior literature has identified four key challenges:
\begin{enumerate}
    \item \textbf{Cognitive ability}: Older adults can struggle mentally with articulating their thoughts and feelings or with engaging in creative thinking, which can limit their ability to contribute to discussions about design ideas such as how they envision future technology could fit into their life \cite{lindsay2012engaging};
    \item \textbf{Physical ability}: Older adults can be physically unable to participate in study activities (\textit{e.g.}, due to physical disability \cite{rogers2022maximizing} or inability to reach study sites \cite{duque2019systematic}), which can lead to certain populations being left out or opting out of participation, limiting representation in design work;
    \item \textbf{Ecosystem}: Older adults can live in complex environments that include customization of the physical space (\textit{e.g.}, ramps, railings, lifts), rigid day-to-day routines and behavioral needs, other individuals who share the space (\textit{e.g.}, family, caregivers), requiring the design process to take into account the entire ecosystems to reach solutions that are acceptable and usable to all stakeholders \cite{gronvall2013participatory};
    \item \textbf{Other stakeholders}: Older adults may no longer be independent in performing activities of daily living (ADL) and rely on people (\textit{e.g.}, family, caregivers) for support, whose needs, constraints, and preferences must also be considered in the design process \cite{hwang2012using}.
\end{enumerate}
Recent research has addressed some of these challenges, particularly to help older adults better grasp the capabilities and limitations of new technology, through the use of higher-fidelity systems \cite{bradwell2021user,sabanovic2015robot}. However, introduction of the technology in a workshop setting may not be sufficient to capture the necessary ecological considerations and the needs of other stakeholders.
We propose \methodfull{} (\method{}), a PD method including elements of user-centered design that addresses some challenges of conducting PD with vulnerable populations as well as design problems where immersion in the use setting is critical to the design process. \method{} situates the activity in a genuine environment, grounds co-design activities in existing technical capabilities or capabilities that can be simulated for participants, centers design activities around experiencing the interaction (as opposed to imagining interactions), and engages other decision makers in the design process. We use this approach to create an immersive, realistic, and reflective co-design experience. The three-phase method, shown in Figure \ref{fig:teaser}, integrates ideas from in-the-wild Wizard of Oz (WoZ) studies \cite{mitchell2021curtain}, user enactments \cite{odom2012fieldwork}, stakeholder involvement \cite{vink2008defining}, and traditional PD workflows. 
\method{} is not disjoint from PD but represents a carefully selected combination of study activities that can facilitate engagement for older adults by considering their cognitive and physical abilities and can help capture the ecological considerations and other stakeholder needs for assistive technologies necessary for successful acceptance and deployment.

We applied \method{} at a senior living facility to design interactions between residents and an assistive mobile robot. Our use of \method{} revealed insights that point to its benefits and limitations. 
Multiple interactions between participants and the robot uncovered significant differences in what people initially designed compared to what they preferred when the robot was performing the scenario. We report on our findings and discuss the benefits of \method{}.
    
Our work makes the following contributions:
\begin{enumerate}
    \item We describe \method{}, a PD method that incorporates realistic, \textit{in situ} interactions throughout the PD process to addresses challenges of designing technologies for older adults;
    \item We employ \method{} in a case study with residents and caregivers of a senior living community to design interactions with an assistive mobile robot;
    \item We present findings from the case study, including insights that reveal the benefits and limitations of \method{};
    \item We discuss \method{}, including its benefits and how it applies to other domains and technologies.
\end{enumerate}

\section{Related Work}
Below, we discuss prior work that informs the development of \method{}.

\subsection{Participatory Design with Older Adults}

Participatory design (PD) has a rich history in human-computer interaction (HCI) to involve stakeholders in the design process. Typical PD activities include watching/discussing videos, creating/considering storyboard scenarios, drawing/sketching ideas, or creating/interacting with low-fidelity prototypes (\textit{e.g.}, paper prototypes) \cite{duque2019systematic}.
The range of technology targeted through PD methods varies widely, including applications that focus on fall prevention \cite{gronvall2013participatory}, mobile communication devices that connect to TVs \cite{scandurra2013participatory}, new banking technologies \cite{vines2012questionable}, and systems that promote healthy eating \cite{lindsay2012engaging}, personal mobility \cite{lindsay2012engaging}, feelings of personal security at home \cite{lindsay2012engaging}, and health tracking \cite{davidson2013health}. 

The human-robot interaction (HRI) community has begun adopting PD methods with older adults, exploring a wide range of robotic designs such as a social robot to help older adults with depression \cite{lee2017steps}, a social robot that hosts GUI-based games for mood stabilization \cite{gasteiger2022participatory}, a mobile robot to reduce falls \cite{eftring2016designing}, and a drink delivery robot \cite{bedaf2017preferred}. Other work, such as that of \citet{broadbent2009retirement} and \citet{bradwell2021user}, focuses on designing how a robot should appear and selecting what tasks are desirable for a robot to complete. Most of these studies do not include the actual robot, and they instead rely on video demonstrations \cite{gasteiger2022participatory, beer2012domesticated}, storyboard images showing what a robot may do \cite{bedaf2017preferred}, or other images of robots \cite{broadbent2009retirement}. While these approaches allow for quick, low-barrier design, the simplicity of the prototypes can make it hard for participants to understand the capability and potential of the artifact, the context of its usage, and how it could fit in their living space. 

Incorporating robots in all design phases has many clear benefits, although the precise use of the robot in PD varies greatly in previous work. In some cases \cite[e.g.,][]{lee2017steps,bradwell2021user,ostrowski2021long}, prototypes are introduced to participants prior to the design session to enhance their understanding of the robot's capabilities.
\citet{ostrowski2021long} also included the robot prototype in the design session itself, but none of these examples conducted any validation with the participants during or after the design process. While this approach seems effective for designing stationary social robots, designing mobile robots, such as some assistive robots, necessitates consideration for the holistic interaction environment.
For example, \citet{eftring2016designing} used PD to design an in-home robot to reduce falls, but never introduced the real robot into the environment until a follow-up field evaluation. Their evaluation found that the robot was too big for some spaces, and participants did not like adding ramps that the robot needed to cross over floor thresholds.
Increasing the use of robots through all phases of the PD process could be critical to developing successful assistive robots with older adults.


\subsection{Other Approaches to Technology Design}
In addition to PD, we can also take inspiration from alternative design methods that offer some insights about how to address challenges of designing assistive robots with older adults:

First, \textit{living labs} emphasize the importance of the context where a technology will be used. By using a study environment that mimics real conditions, researchers can understand how a technology will function in that space \cite{alavi2020five}. However, living labs do not emphasize engaging stakeholders as strongly as methods such as PD \cite{bygholm2017not}.

Second, \textit{Wizard of Oz} (WoZ) allows participants to interact with a system that is controlled by an operator behind the scenes \cite{dahlback1993wizard}. It has been used in laboratory settings to design interfaces and system behavior through tools such as Ozlab \cite{pettersson2002ozlab,larsson2006rapid,wik2020wizardry}. \citet{mitchell2021curtain} discusses the need for in-the-wild WoZ studies to capture more natural interactions that reveal usability challenges that would otherwise be missed, but they focus their use of WoZ for system evaluation instead of during the design process.

Third, \textit{role playing} has been used to engage potential end users in the design of future technology \cite{odom2012fieldwork,stromberg2004interactive}. \citet{odom2012fieldwork} specifically discusses how \textit{user enactments} (UE) can allow researchers to quickly explore how technology fits into an environment. 
While these methods facilitate good participant engagement, the staged setups and lack of usable prototypes limit the ability of participants to experience the technology as they would in their daily life.


\subsection{The Special Case of HRI in Assisted Living}

Assisted living is a type of senior living community for individuals who are no longer able to live independently \cite{zimmerman2007definition}. Residents typically live in private rooms with shared dining halls and other common spaces, placing this living arrangement somewhere in between a private residence and a more clinical setting such as a hospital or skilled nursing facility. Throughout the day, residents in assisted living can expect to receive regular help from caregivers for activities necessary for living independently, which can include care tasks such as bathing, dressing, toileting, transferring to or from a bed or chair, laundry, and more \cite{kane1993assisted}. They may also receive light medical assistance, primarily in the form of physical or occupational therapy and medication management \cite{kane1993assisted}.

Technology for assisted living settings aims to enhance the livelihood and independence of the residents and also ease the burden of caregivers. For example, ambient assisted living (AAL) incorporates smart home technology into living spaces to improve the safety, health, and well-being of residents \cite{aced2015supporting,zulas2012caregiver}. Socially assistive robots are being developed for applications such as providing health reminders and assisting older adults to manage symptoms of depression \cite{bradwell2021user}. Assistive robots are being explored to perform tasks such as refilling water \cite{odabasi2022refilling}, helping with ambulation \cite{mederic2004design}, and escorting residents to activities \cite{pollack2002pearl}.
The technology being used day to day in assisted living settings is also modernizing. For example, we have already seen vacuum robots and computerized medication dispensing carts commercially deployed in care facilities. 

Despite research advances and industry adoption of new technology, it is not yet clear how assistive robots should fit. To better incorporate robots in care settings, \citet{bardaro2021robots} and \citet{hornecker2020interactive} recommend working with a variety of stakeholders to identify specific needs that robots can address. \citet{stegner2022designing} and 
\citet{alaiad2014determinants} build on this work by identifying complex and potentially conflicting power dynamics in care settings. 

As robotic systems are developed, it is critical to consider them in a broader context, such as how the robot will come and go between private and public spaces in the facility, who assigns tasks to the robot, and how to balance caregiver and resident preferences with regard to robot behaviors.
However, current design approaches for assistive robots with older adults primarily focus on details such as robot appearance, technical performance, or overall acceptance of the robot. Instead, we need to think about how robots fit more holistically into the assisted living setting. To help address these open questions, we can take lessons from HCI design methods 
and apply them to HRI with older adults. Specifically, we consider how situated interactions with technology could be used to overcome established challenges of using PD with older adults and understand some aspects of system deployability in real-world environments.
\section{Research Questions}
\label{rq}
To successfully relieve caregiver burden and increase resident independence, assistive robots need to address real needs within senior living communities. Robotic systems need to be sufficiently capable, but they also need to meet the expectations residents have regarding how the system can fit into their day-to-day activities and need to be compatible with how caregivers provide care to residents. 
Motivated by these needs and the challenges identified in \S\ref{intro}, we pose the following research questions: 

\textbf{RQ1}: How can designers effectively engage older adults to better contribute to the design of assistive technologies?

\textbf{RQ2}: How can designers better understand the challenges of integrating assistive technologies in genuine environments, interactions, life activities, and caregiving practices for older adults?


This work explores the research questions proposed above with a focus on robotic systems. Our intuition to answer these questions is that situating design ideas directly in the real environment can provide us with the insights needed. 

\section{Method}
The previous work on PD with older adults guided us in crafting \method{}. 
In this section, we first discuss our research context, including an overview of \method{}, case study goal, community partner, participants, and robotic platform. Then, we present the key phases of \method{} by describing the general concept of each phase and presenting their application in a case study at a senior living facility.

\begin{figure*}[!b]
    \centering
    \includegraphics[width=\linewidth]{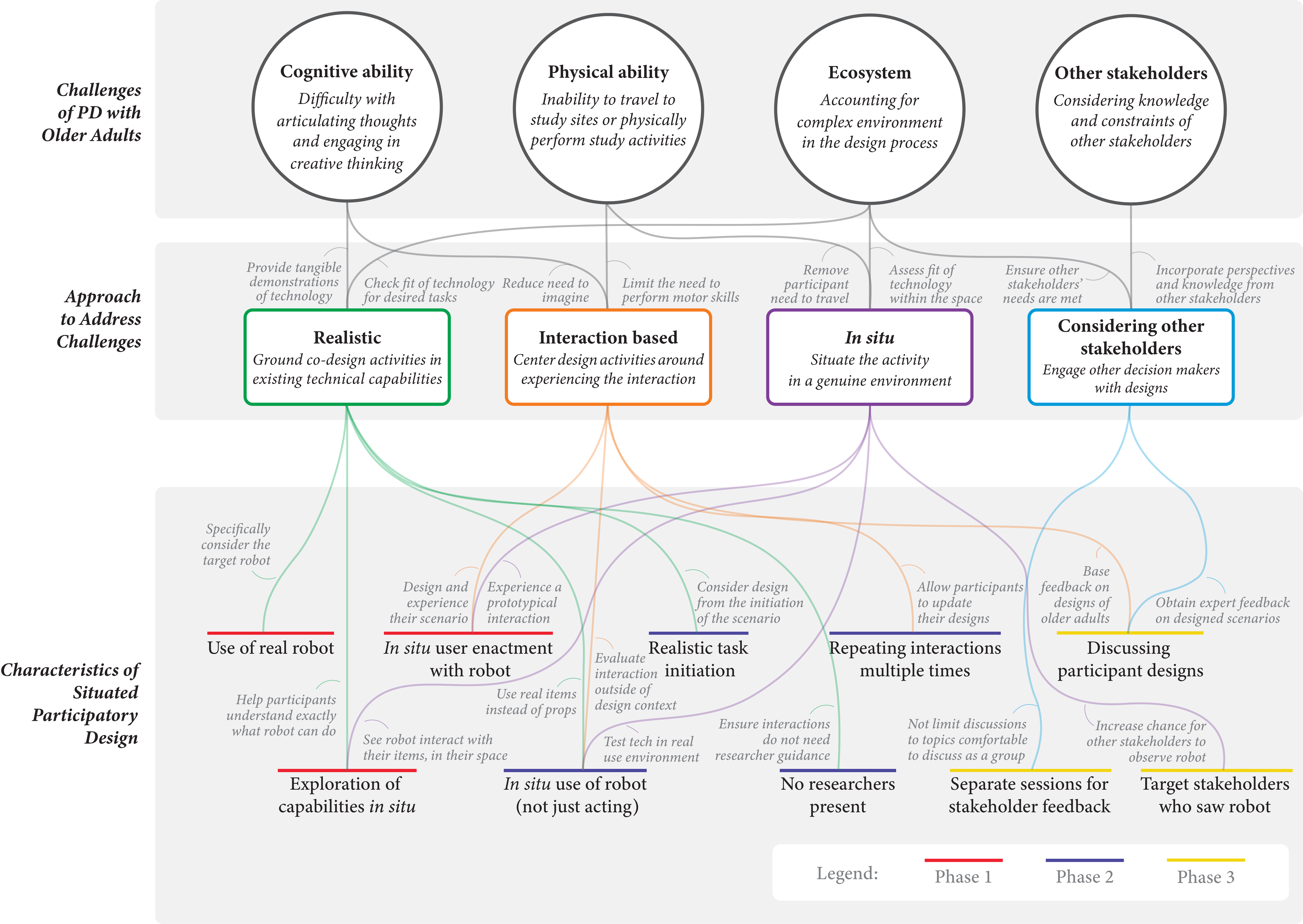}
    \caption{Conceptual development of \method{} from the challenges of PD with older adults to our three-phase method. Motivated by the challenges of PD with older adults identified from previous literature (top row, see \S\ref{intro} for more details), we identified a general approach to addressing these challenges through the integration of user-centered design approaches (middle row), formulating key characteristics for \method{} that instantiate these approaches (bottom row).}
    \Description{Graph with three rows with the headings: Challenges of PD with older adults, Approach to address challenges, and Elements of Situated Participatory Design.

The first row has 4 circles with the texts:
1. Cognitive ability: difficulty with articulating thoughts and engaging in creative thinking
2. Physical ability: Inability to travel to study site or physically perform study activities
3. Ecosystem: Accounting for complex environment in the design process
4. Other stakeholders: Considering knowledge and constraints of other stakeholders

The second row has 4 boxes with the texts:
1. Realistic: Ground co-design activities in existing technical capabilities
2.Interaction based: Center design activities around experiencing the interaction
3. In situ: Situate the activity in a genuine environment
4. Considering other stakeholders: engage other decision makers with design

With connections from circles in 1 to boxes in 2:
1 to 1: Provide tangible demonstrations of technology
1 to 2: Reduce need to imagine
2 to 2: Limit the need to perform motor skills
2 to 3: Remove participant need to travel
3 to 1: Check fit of technology for desired tasks
3 to 3: Assess fit of technology within the space
3 to 4: Ensure other stakeholder’ needs are met
4 to 4: Incorporate perspectives & knowledge from other stakeholders

The third row has 10 labels divided in three phases:
Phase 1:
1. Use of real robot
2. Exploration of capabilities in situ
3. In situ user enactment with robot 

Phase 2:
4. In situ use of robot (not just acting)
5. Realistic task initiation
6. No researchers present
7. Repeating interactions multiple times

Phase 3:
8. Separate sessions for stakeholder feedback
9. Discussing Participant designs
10. Target Stakeholders who saw robot

With connections from boxes in row 2 to labels in row 3:
1 to 1: Specifically consider the target robot
1 to 2: Help participants understand exactly what the robot can do
1 to 4: Use real items instead of props
1 to 5: Consider design when it starts, without prompting
1 to 6: Ensure interactions do not need researcher guidance
2 to 3: Design and experience their interaction
2 to 4: Evaluate interaction outside of design context
2 to 7: Allow participants to update their designs
2 to 9: Base feedback on designs of older adults
3 to 2: See robot interact with their items, in their space
3 to 3: Experience a prototypical interaction
3 to 4: Test tech in real use environment
3 to 10: Increase chance for other stakeholders to observe robot
4 to 8: Not limit discussion to topics comfortable to discuss as a group
4 to 9: Obtain expert feedback on design scenario}
    \label{fig:method_dev}
\end{figure*}

\subsection{\method{} Overview}
\method{} is an iterative approach to designing technology when the goal is an eventual deployment. 
We developed \method{} based on the challenges we identified in \S\ref{intro} for PD with older adults relating to cognitive ability, physical ability, ecosystem, and other stakeholders. To address these challenges, we devised an approach that integrates situating the activity in a genuine environment, grounding co-design activities in existing technical capabilities, centering design activities around experiencing the interaction, and engaging other decision makers in the design process. This approach provides the foundation for the following three-phase method:
\begin{itemize}
    \item \textit{Phase 1: Discovery, co-design, \& enactment} --- use the real technology \textit{in situ} to explore its capabilities as well as select, design, and enact scenarios;
    \item \textit{Phase 2: Simulated deployments} --- evaluate the designed scenarios multiple times under realistic conditions using in-the-wild Wizard of Oz (WoZ) (\textit{i.e.}, \textit{in situ} use of the real robot with the participant's real belongings, realistic task initiation, and without the researchers present to mediate);
    \item \textit{Phase 3: Engaging other stakeholders} --- conduct separate sessions with other stakeholders (\textit{e.g.,} caregivers) to present participant designs and discuss experiences and concerns.
\end{itemize} 
The evolution from identifying the challenges to formulating characteristics for \method{} is detailed in Figure \ref{fig:method_dev}.
Each phase builds upon the previously gained knowledge, and this design cycle could be repeated until the design reaches the desired level of maturity.

\begin{figure}[!tb]
    \centering
    \includegraphics[width=\columnwidth]{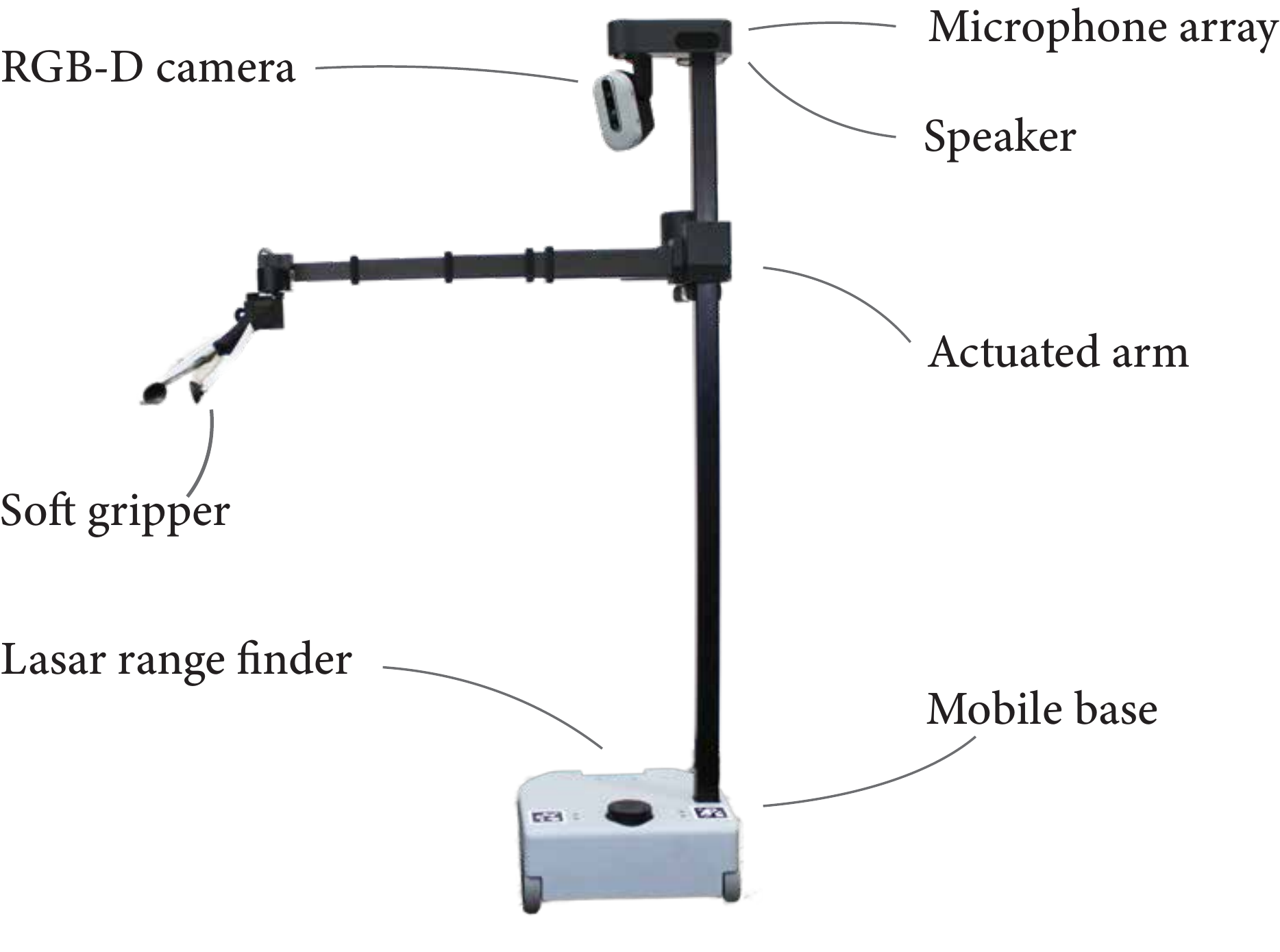}
    \caption{Stretch RE1 mobile manipulator robot from Hello Robot. We used Stretch in our case study with older adults.}
    \Description{Annotated picture of the stretch robot. The robot has a mobile base with two wheels and laser range finders and the non-anthropomorphic ``head'' of the robot is a RGB-D camera on a pan-tilt mount with a microphone array and a speaker. The ``head'' and the base are connected by a rectangular pole along which a telescopic arm can go up and down. The end of the arm is a soft gripper looking similar to tongs.}
    \label{fig:stretch}
\end{figure}

\subsection{Case Study Details}
\subsubsection{Case Study Goal}
Our case study builds on work by \citet{stegner2022designing}, which offers insights into the day-to-day practices of professional caregivers and the needs of older adults living in assisted and independent living facilities. We  use \method{} to investigate residents' perspectives on how a robot could fit into their daily lives by specifically focusing on light manipulation tasks such as delivering a cup of water or picking an item up from the floor. 

\subsubsection{Community Partner}
We partnered with a suburban, private, not-for-profit senior living facility located in the Midwestern United States. The facility includes a mixture of accommodations, including 60 Assisted Living (AL) apartments and 85 Independent Living (IL) apartments. We primarily worked with AL residents, as this population could benefit significantly from light manipulation assistance, but we also involved IL residents who expressed interest. Most residents in IL are completely independent, but some receive assistance with medication management or other light tasks such as bathing or getting dressed. 
Similarly to other care facilities, our community partner has faced recent difficulty with staffing and are frequently understaffed or staffed with temporary workers.

\subsubsection{Participants}
In total, nine residents, aged 77--94 years ($M=88.3$ years, $SD=5.8$ years; 6 females; 7 in AL, 2 in IL), participated in the study. We do not report individual characteristics to minimize any risk of re-identification given the small population from which we sampled. However, we can report that many of our participants had mobility, dexterity, visual, or hearing impairments.
Participants received \$20 USD/hour to participate in Phase 1 and a flat fee of \$20 USD to participate in Phase 2. Our community partner helped recruit participants who expressed interest and who were directly able to provide informed consent to participate.

In addition, three caregivers participated in Phase 3, aged 22--54 years ($M=33.3$ years, $SD=14.3$ years; all female) with experience varying between 6 months to 5 years ($M=2.5$ years, $SD=1.9$ years). 
Each interview lasted 30 minutes, and caregivers were compensated at a rate of \$40 USD/hour for their time.

\subsubsection{Robot Platform}
We used the Stretch RE1 robot from Hello Robot \cite{kemp2022design}, shown in Figure \ref{fig:stretch}, as our robot platform. Stretch is a mobile collaborative robot (cobot) that is 55.5 inches, or 141 cm, tall and equipped with a laser range finder, RGB-D camera, microphone array, speaker, and actuated arm with a soft gripper that can lift up to 3.3 lbs, or 1.5 kg. Throughout the design sessions, we realized that the base capabilities of Stretch were too limited for our use case (\textit{e.g.}, the speakers were not loud enough; the onboard camera was not sufficient for remote operation), and thus we augmented the Stretch robot with three additional cameras and a Bluetooth speaker to conduct the study. The robot's remote operation was conducted through a mixture of a gamepad controller using the default Stretch teleoperation software\footnote{Stretch teleoperation software: \url{https://github.com/hello-robot/stretch_body/blob/master/tools/bin/stretch_xbox_controller_teleop.py}} and a dedicated web app for displaying camera feed and typing sentences for the robot to speak. We initially used the default Google Assistant Red voice, but based on participant feedback during the study we switched to use Amazon Polly with the Joey voice slowed down to 70\% as our text-to-speech platform for the robot's prompts and responses.

\subsection{Procedure}

Applying one cycle of \method{}, we conducted a field study during Summer 2022 at our community partner facility to explore the design space of robot-assisted care activities for older adults. All study methods were reviewed and approved by our institutional review board (IRB). Study materials and results are provided via OSF.\footnote{Study data and materials are available on \textit{OSF}: \url{https://osf.io/ubnw5/}}

We present the general phase description in parallel with the steps of our case study to illustrate how \method{} can be applied to a real-world design scenario. We will refer to the facilitators of the design session, \textit{researchers}, and users who took part in the session, \textit{participants} (\textit{residents} in Phases 1 and 2, and \textit{caregivers} in Phase 3).

\begin{figure*}[!t]
    \centering
    \includegraphics[width=\textwidth]{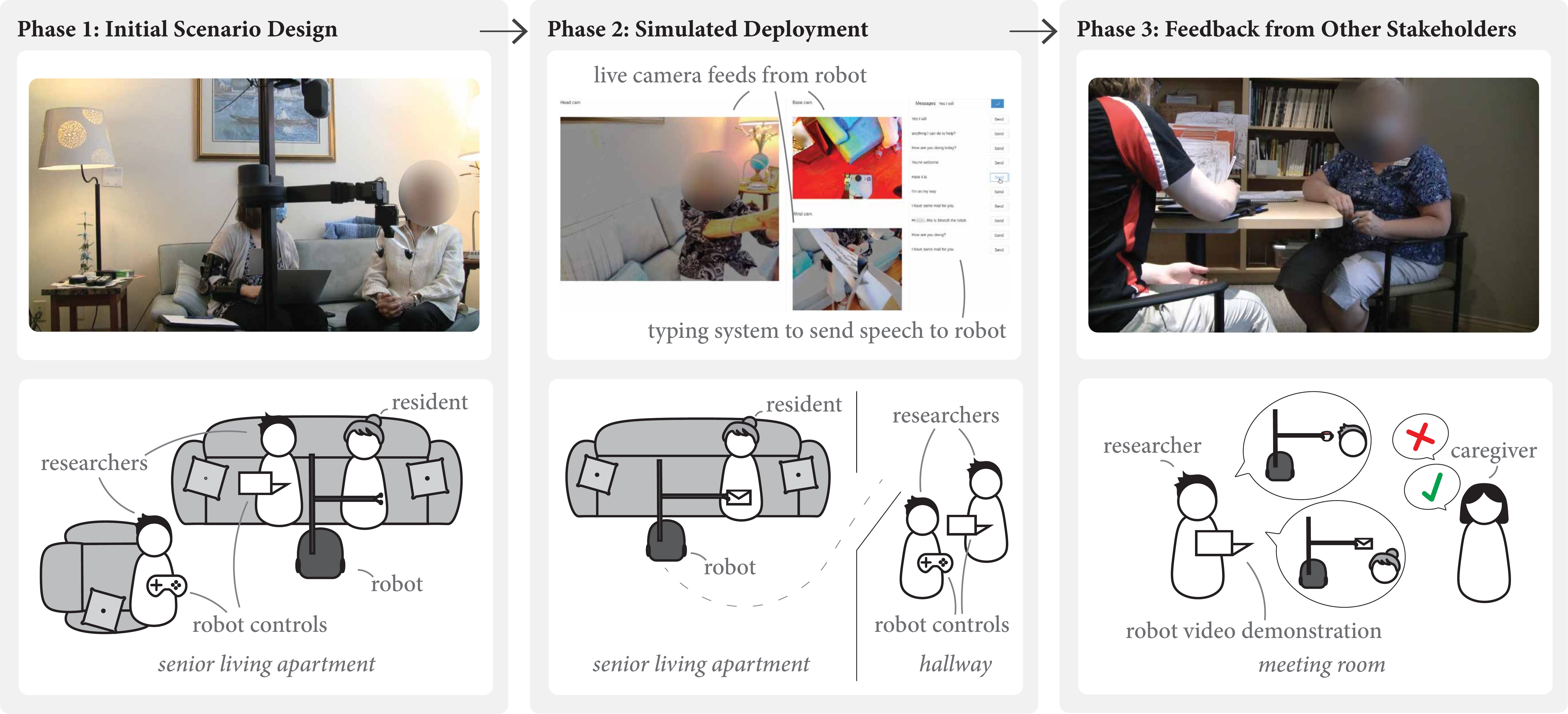}
    \caption{Each phase of \method{} illustrated with pictures from our case study. The first phase involves a \textit{co-design activity} where researchers work with the older adult in their home to design a scenario (left). The second phase involves \textit{simulated deployment}, where the researchers remotely operate a robot using the web app shown and a gamepad controller to complete the scenario with the resident (center). The third phase involves \textit{follow-up interviews} with caregivers at the facility to reflect on resident designs within the context of their care practices and address aspects of the scenarios that are specific to caregivers (right).}
    \Description{Example of the method through six panels describing the three steps with pictures from the study and the same simple sketches used in Figure 1. 
  The first column titled ``Phase 1: Initial Scenario Design''. 
  The top image shows a picture of a researchers and an older adult on a could with the stretch robot. The bottom image shows a sketch with two researchers, an older adult sitting together in a resident home and a robot in the middle. The two researchers control the robot which interacts with the older adult. 
  The second column is titled ``Phase 2: Simulated Deployment''.
  The top image shows an annotated screenshot of the interface with three rectangle showing live views from the three robot camera (head, base, and wrist) and a panel on the right with predefined sentences such as ``You're welcome'' with a ``sending'' button and an open field. The bottom image shows the robot interacting with the  older adult, while the researchers are hidden in the hallway to remotely operate the robot.
  The third column is titled ``Phase 3: Feedback from Other Stakeholders''. 
  The top images shows a picture of the researcher and a caregiver in a library. The bottom image is a sketch showing an researcher talking to a caregiver in a meeting room and showing them videos of interactions between residents and the robot on a laptop and the caregiver providing feedback on the interactions.}
    \label{fig:case_study}
\end{figure*}

\subsubsection{Phase 1: Discovery, Co-design, \& Enactment}
\paragraph{Description.  }
Phase 1 combines concepts from PD and user enactment. The researchers first introduce participants to the goal of the research and gain an understanding of that participant's individual needs and circumstances. Then, the technology is introduced and participants interact with it based on a scenario that is personally relevant to them. This activity provides an initial scenario design that will be used and modified throughout the rest of the study. Once the initial design is set, the researchers facilitate user enactments, where a researcher remotely operates the technology to allow the participant to walk through their design and provide feedback. Researchers should focus the scenario design based on reasonable capabilities of the technology, although they may have to intervene in instances that the current prototype is not yet able to execute (\textit{e.g.}, opening a door to let the robot in).

\paragraph{Case Study Application.  }
Phase 1 consisted of a single hour-long session per participant. The key elements in Phase 1 include:
\begin{enumerate}
    \item \textit{Ice breaking \& Needfinding}: We started by introducing participants to the goal of the research and the plan for the study. We then asked them to describe their typical day and with which tasks they typically receive assistance. For each task, we noted on a card the type of activity, frequency (how often the resident needs help with it), timing (when do they often need the assistance), scheduling (whether it is planned or unplanned), initiation (who prompts the task to start), and comfort (would they be comfortable with a robot providing this assistance). During this time, the robot was out of the room to avoid distraction, and as the participants had yet to see the robot, responses were mostly \textit{a priori}.
    \item \textit{Robot Introduction}: We brought the robot into the room and gave a demonstration and verbal description of its abilities. During this step (see Figure \ref{fig:case_study}, left), we controlled the robot in full view of the participants, describing to them how we could move parts or make the robot speak. As we demonstrated each feature, the residents had the opportunity to interact with the robot and ask questions about it or its capabilities.
    \item \textit{Interaction Design}: From the tasks that the resident provided earlier, the researchers selected a task for the robot to do based on a combination of the robot's capabilities and the resident's interest in what a robot should do. Once the task was agreed upon, we used it as a prompt to design the scenario together. As a grounding point, the resident described what steps the caregiver would normally do to complete the task. These steps were recorded on a worksheet. Then, we asked the resident to consider if our robot was doing the task, how should its behavior change.
    \item \textit{Enactment}: Based on the resident's initial design, we used the robot to enact the scenario with the resident. During the enactment, researchers were next to the resident and the resident had the opportunity to request changes or provide feedback. In a brief follow-up, the resident answered questions about their experience, \textit{e.g.}, whether the interaction met their expectations and if any changes should be made.
\end{enumerate}
After Phase 1, we arranged for the robot to return for Phase 2 to validate the design through two simulated deployment sessions.

\subsubsection{Phase 2: Simulated Deployments.}
\paragraph{Description.  }
Phase 2 integrates in-the-wild Wizard of Oz (WoZ) \cite{mitchell2021curtain} through multiple, iterative sessions where researchers simulate the deployment of the technology in a way that reflects the participant's design. The simulated deployment provides the opportunity to uncover ecological considerations that are important to consider for future deployments. Participants are asked to simply use the technology as they had co-designed in Phase 1, and the interaction is completed as realistically as possible. We create the realistic context by using real items instead of props when possible, matching the time to when the participant would typically engage in the scenario, and removing the researchers from mediating the interaction.
After the simulated deployment in a short interview with the researchers, the participant is asked to reflect and give feedback on their experience as input into an updated design. 

\paragraph{Case Study Application.  }
In Phase 2, we held two sessions lasting approximately 15 minutes each and occurring on different days. The key elements for one single session of Phase 2 were as follows:
\begin{enumerate}
    \item \textit{Simulated Deployment}: Based on the scenario design that resulted from Phase 1, we controlled the robot through WoZ to have the robot enter the resident's room and complete the scenario. Two researchers who are out of sight of the resident operated the robot: one researcher controlled the robot's movement using a gamepad controller, and the other controlled the robot's speech. Figure \ref{fig:case_study} (center) shows the interface used to stream cameras to assist in remote operation and send speech for the robot to say. The original setup included streaming the microphone data from the robot, but the microphone did not reliably capture participant speech, so the researchers listened through the door. 
    \item \textit{Reflective Interviews}: After the first simulated deployment, we briefly (3--5 minutes) interviewed the participant about their experience with the robot and gave them the opportunity to propose changes. After the second simulated deployment, we conducted longer (10--15 minutes) interviews to probe into additional wider-ranging questions such as, ``After having experienced the interaction with the robot, would you prefer a human or robot to perform the task?'' and  ``Do you see yourself using a service like this in your daily life?''
\end{enumerate}

\subsubsection{Phase 3: Engaging Other Stakeholders.}
\paragraph{Description.  }
Phase 3 is a follow up to Phase 1 and Phase 2, based on the concept of expert feedback. Since the direct users are not the only stakeholders in the interaction, it is critical to also involve other stakeholders. For example, in the case of assisted living, older adults rely on formal and informal caregivers to provide assistance. 
This phase seeks to gather feedback on whether the designs of the participants are reasonable and safe and other considerations that may not emerge from working directly with the target users. Whereas the focus of Phase 1 and Phase 2 is a scenario specific to an individual participant, Phase 3 allows other stakeholders to provide input on multiple participants' scenarios at once. This phase also provides an opportunity to resolve sensitive and controversial design decisions, such as features where a participant and an expert might disagree (\textit{e.g.}, a nutritionist recommending minimizing sugar versus the client wanting sweet snacks to be delivered by technology). These insights can fill in missing facets of the design without adding tension between the participants and other stakeholders.

\paragraph{Case Study Application.  }
After completing Phase 2, we interviewed (approximately 30 minutes each) caregivers at the facility. Due to a COVID-19 outbreak, our data collection with the caregivers was shorter than planned. Sessions were conducted in person or through a Zoom video call. Although we aimed to recruit caregivers who had previously seen the robot during Phase 1 or Phase 2 while we worked with the robot with the residents, in practice, staffing challenges at the facility made this approach infeasible. Instead, we recruited caregivers who regularly worked at the facility, as opposed to temporary workers used to cover staffing shortages.

During the interviews, shown in Figure \ref{fig:case_study} (right), the researchers gave an overview of our research aim and asked the caregiver to reflect on their knowledge of the robot, including anything they heard from residents or other staff. Then, the researchers presented the scenarios designed by the residents and asked for their feedback. Finally, the caregivers provided input on key design decisions that they were uniquely positioned to consider, such as who should personalize the robot to each resident's preferences and how much oversight  caregivers should have over the robot. 

\subsection{Data Collection \modified{\& Analysis}}

We collected three forms of data throughout the study: researcher field notes throughout the various study sessions (\textit{i.e.}, activity cards from Phase 1 and notes from interviews in Phase 2 and Phase 3), participant-generated designs, and video/audio recordings during design sessions and interviews. Since the design sessions are highly contextualized in the real-world environment, we did not transcribe the audio/video data but instead used a bookmarking system where researchers marked points of interest within the field notes to allow quick access to revisit key moments in the video/audio data.

Data was analyzed using a Reflexive Thematic Analysis approach \cite{braun2022conceptual}. The two researchers who conducted the study sessions, who were already familiar with the data, performed the analysis. The first author used open-coding to identify phenomena from the field notes and participant designs, revisiting the recordings as necessary for context. The two researchers then worked together to discuss and refine the codes, following an iterative approach to organize the codes into insights using affinity diagramming.

During the open coding and affinity diagramming, we focused on two high-level ideas in the data. First, we considered the design findings from participants to inform future robot design and deployments. Second, we considered the data as it pertains to \method{} in order to identify insights we gained from using the method. In this paper, we emphasize the methodological findings and provide only a summary of the findings on robot design, which we still think is informative to understand the benefits and limitations of \method{}.

\section{Findings}
We present the results from our case study, organized into two sections: (1) design findings from participants to inform future robot design and deployments, (2) insights into \method{} that emerged from the case study. 
Findings are supported by researcher observations and quotes from participants.
Both quotes and observations are attributed using participant ID, with residents as R1--R9 and caregivers as C1--C3. We made minimal edits and added annotations to the quotes to improve their clarity while retaining their meaning.


\begin{table*}[!t]
\caption{A selection of scenarios that participants designed for the robot, including significant features of their envisioned interaction with the robot. Each participant selected a task that was relevant to their day-to-day activities and needs. While their designs evolved throughout the study, this snapshot represents their resulting designs at the end of Phase 2.}
\label{tab:participant_designs}
\centering\small
\begin{tabular}{ p{0.18\linewidth}p{0.18\linewidth}p{0.18\linewidth}p{0.18\linewidth}p{0.18\linewidth} }
    \toprule
     & \textbf{R1} & \textbf{R3} & \textbf{R6} & \textbf{R8} \\ 
    \midrule
    \textit{Task} & Water bottle delivery & Mail delivery & Move cup of water & Cup of ice delivery \\
    \midrule
    \textit{How is the task initiated?} & Pre-arranged times, or on-demand calls. & Brought when it arrives. & R6 wanted to press a button to call robot. & Pre-arranged time\newline (4 pm sharp). \\
    \addlinespace[.2cm]
    \textit{How should the robot enter?} & Knock, wait for a response; Key needed to enter. & Knock/make announcement, enter without waiting. & Knock, wait to enter. & If the door is open, enter;\newline else, knock and enter. \\
    \addlinespace[.2cm]
    \textit{How should the task be\newline completed?} & Retrieve the water bottle from refrigerator and set it on the side table. & Bring the mail to R3 wherever they are. & R6 will give the robot\newline specific instructions. & If R8 is in the room, bring it to them; otherwise leave it on the side table. \\
    \addlinespace[.2cm]
    \textit{What other behavior from the robot is desirable?} & Light conversation; \newline Prior to leaving, schedule next task. & Voice updates on robot progress; \newline Minimal, polite speech. & Complex conversation; \newline Offer to do anything else\newline before leaving. & Little bit of speech. \\
    \bottomrule
\end{tabular}
\end{table*}

\subsection{Participant Designs and Feedback}
Below, we overview the scenarios designed by participants and the design findings based on feedback from participants.

\subsubsection{Scenarios.} Participants designed scenarios for a wide range of tasks for the robot, including mail, newspaper, book, or water bottle delivery; refilling ice water; moving a cup of water across the room; and picking items up from the floor. As Phase 1 and Phase 2 progressed, design ideas evolved based on participant experience (see \S\ref{phase2}). 
Table \ref{tab:participant_designs} summarizes sample interactions, including the scenario and key behavioral expectations from the robot. 

\subsubsection{Feedback.} Our analysis resulted in themes on the behavioral expectations for, physical attributes of, interaction quality with, and attitudes toward the robot. The range of preferences supports other work calling for personalized robots and similar systems.

\textit{Behavioral expectations ---} Behavioral expectations included preferences on the socialness of the robot; some residents desired a highly conversational agent, while others wanted the task to be completed in silence. Other behavioral expectations included how the robot should gain entry into the resident's space: knock and wait for a response, knock and enter without waiting, or directly enter without warning if the door is open. In some cases, the residents also kept their doors locked, so the robot would additionally need a key to gain access. Several residents also expressed concerns over the how the robot would interact with their personal belongings, which limited the tasks they felt appropriate for the robot to complete. Specific concerns included ``security of [the robot having] the mail'' (R2) or that the robot would ``spill'' (R6) something.

\textit{Physical attributes ---} For physical attributes, participants commented on the size of the robot, movement speed, the robot's voice, and the timing of speech. While some participants appreciated the small form factor of the robot, one participant in a wheelchair remarked they ``didn't think they could communicate'' because the robot was too tall.
Participants also perceived the robot's movement as ``slow'' (R1), which impacted some of their future preferences. 

\textit{Interaction quality ---} With interaction quality, the robot's speech was the main factor. We found that the initial style and volume of the robot's voice were too quiet for residents to ``understand the words'' (R8) and too ``high-pitched'' (R2) for them to hear, which is what prompted us to change the text-to-speech (TTS) engine and add an external speaker. The timing of the robot's speech during conversations with participants was also challenging. 
Participants struggled to understand when the robot ``paused'' (R5) before speaking. Some of them suggested that the robot should provide ``a simple [visual] movement'' (R5) to signal its processing state, while others felt it would ``just take time'' (R9) to learn how to ``interact'' (R9).

\textit{Attitude toward the robot ---} Finally, attitude toward the robot encompassed thoughts on whether the participants preferred a human or robot to complete certain tasks. We observed three main categories of participants. Some preferred a human caregiver even after experiencing the robot. Others felt it was ``immaterial'' (R2) whether it was a human or robot, as long as the robot was ``efficient in supplementing human care'' (R3). A few participants felt the robot was more desirable --- they sometimes felt they were being a ``nuisance'' (R8) asking caregiver to help them, while they would be more comfortable asking the robot to do some tasks. 


\begin{table*}[!h]
\caption{Summary of the insights gained on \method{} from the case study mapped on the characteristics of our method.}
\label{tab:results_summary}
\centering\small
\begin{tabular}{p{0.31\linewidth}p{0.31\linewidth}p{0.31\linewidth}} 
    \toprule
    \textbf{Characteristic of Method} & \textbf{Resulting Insight} & \textbf{Example}\\ 
    \midrule
    \textit{Phase 1} \\ 
    \addlinespace[.1cm]
    Use of real robot & I1: Introducing the robot first helps uncover participant comfort. & R4 was initially enthusiastic about the study but then withdrew once they saw the robot. \\[.2cm]
    Exploration of capabilities \textit{in situ} & I2: Exploring robot capabilities directly with residents allows both the residents and the researchers to envision how the robot can address the resident's needs. & R6 prepared tasks for the robot to test its capabilities, giving R6 a better idea of the robot's capabilities and the researchers insight into the robot manipulating items outside of a lab setting. \\[.2cm]
    \textit{In situ} user enactment with robot & I3: Experiencing the interaction is an effective way to explore design decisions. & R5 struggled to imagine how the robot should behave, but through User Enactment was able to realize what they wanted the robot to do. \\[.2cm]
    \midrule
    \textit{Phase 2} \\
    \addlinespace[.1cm]
    Repeating interactions multiple times & I4: Iterative interactions enable reflection on preferences for robot behavior. & R9 initially wanted a simple robot interaction, but through multiple iterations, they realized they wanted more updates from the robot and that it should have deeper conversation abilities. \\[.2cm]
    \textit{In situ} use of the robot (not just acting) & I5: Experiencing the realistic scenario helps participants realize how it fits into their lives. & R8 initially wanted the robot to deliver the morning newspaper, but realized after the first simulated deployment that they felt a different task would be more suitable for the robot to do. \\[.2cm]
    Realistic task initiation & I6: Unexpected situations can appear from experiencing the robot in day-to-day life. & R1 had a scooter blocking the robot's way, forcing a different strategy to deliver the water bottle. \\[.2cm]
    No researchers present & I7: Interacting with the robot without the mediation of the researchers can facilitate problem solving and idea generation. & R7 expressed they were unsure how to interact with the robot and requested the robot provide them with guidance on how to do so. \\[.2cm]
    \midrule
    \textit{Phase 3} \\
    \addlinespace[.1cm]
    Target stakeholders who saw robot & I8: Familiarity with the robot helps shape caregiver expectations for what the robot can do. & C1 emphasized it was helpful to have seen the robot around to give her a better picture of it and imagine what it could do to help the residents.  \\[.2cm]
    Discuss participant designs & I9: Common ground creates an environment where we can get meaningful feedback about the robot. & C3 related to our experience with customizing robot behaviors for each resident and confirmed the need for the robot to interact with different residents based on their needs and preferences.  \\[.2cm]
    Separate sessions for stakeholder feedback & I10: Discussion of the robot's role in assisted living elicits reflection on authority over the robot. & C2 preferred to have oversight of the robot, but mentioned the ethics of protecting resident independence while looking out for their safety. \\[.1cm]
    \bottomrule
\end{tabular}
\end{table*}

\subsection{Insights into \method{} from the Case Study}
Below, we present the insights we gained from interacting with the residents in Phases 1 and 2 and caregivers in Phase 3 that emerged as a result of \method{}. We describe each insight briefly and offer an illustrative example of it from our case study. Table~\ref{tab:results_summary} summarizes the insights and maps them to the various components of \method{}.

\subsubsection{Insights from Engaging with Residents in Phase 1}

\paragraph{I1: Introducing the robot first helps uncover participant comfort.}
The robot was maneuvering in participants' private rooms, sometimes getting very close to them to perform handoffs or similar tasks. The physical presence of the robot elicited differing responses. 

R4 withdrew from the study because the robot made them uncomfortable. When initially discussing the concept of an assistive robot, R4 was attentive and curious, and even smiled when the robot first entered. However, as the robot was moving around and interacting with R4, their demeanor changed, and they became too distressed from the robot's presence to continue with the study.

Despite this unique example, most participants were comfortable in the presence of the robot, even when it entered into close proximity such as to complete a hand off (\textit{e.g.}, deliver the newspaper). R8 expressed that they were ``very comfortable'' with the robot approaching them and that they were ``confident that he [the robot] was going to stop and [...] not run into me or push me over.'' 

Varying reactions to the robot's presence shows how bringing the robot early in the design process is key to evaluate early on whether the robot could be acceptable.



\paragraph{I2: Exploring robot capabilities directly with residents allows both the residents and the researchers to envision how the robot can address the resident's needs.}
Since our setup allowed real-time control of the robot, participants had ample time to explore the robot's capabilities. While some residents were content to view a demonstration of the system and verbally ask questions about it, others wanted to see if the robot could do specific tasks that they envisioned. We tried every task participants asked us to try, which gave them a chance to witness the robot's abilities and us a chance to assess the challenges of doing a variety of tasks with real items in a real space.

R6 in particular wanted to explore what the robot could do. During the robot introduction in Phase 1, R6 eagerly wanted to test the robot, asking ``Do we try? Shall we try?'' Without prompting, R6 had prepared tasks to ask the robot to try during the session: pick up a tissue from floor, move their cup across the room, unscrew the cap from a nutrition drink, and arrange items of clothing in the closet. From having the robot interact with R6's personal belongings, such as their favorite cup, we gained practical insights into the challenges of the robot grasping and lifting real-world items outside of a laboratory setting. While the robot could complete the first two tasks, it was unable to do the others. R6 was disappointed that the robot could not ``open cans, like water bottles,'' although they were pleased overall with the robot's ability to provide assistance.

\paragraph{I3: Experiencing the interaction is an effective way to explore design decisions.}
During Phase 1, in the initial co-design step, we asked participants how the robot should behave as it completes the agreed-upon scenario. While some residents could articulate an initial version of what the robot should do, not all were able to imagine it. Through the user enactment, they had the opportunity to realize what the robot should do by trying it out.

R5 enjoyed discussing the robot, but expressed difficulty answering questions about what the robot should do at various stages of the interaction. Eventually, they said, ``I'll learn what I want it to do by experiencing it and finding out.'' 
While we were unable to complete the initial co-design activity, we proceeded with the user enactment. Through the enactment, R5 was able to articulate what the robot should do by experiencing the scenario. For example, R5 could not imagine how the robot should behave once it entered the room. During the user enactment, however, they were naturally talking to the robot and giving it instructions on what to do with the mail it was delivering.
This example demonstrates how the enactments helped extract \textit{tacit knowledge} \cite{polanyi2009tacit} from participants that they otherwise struggled to communicate.

\begin{figure*}[!b]
    \centering
    \includegraphics[width=\textwidth]{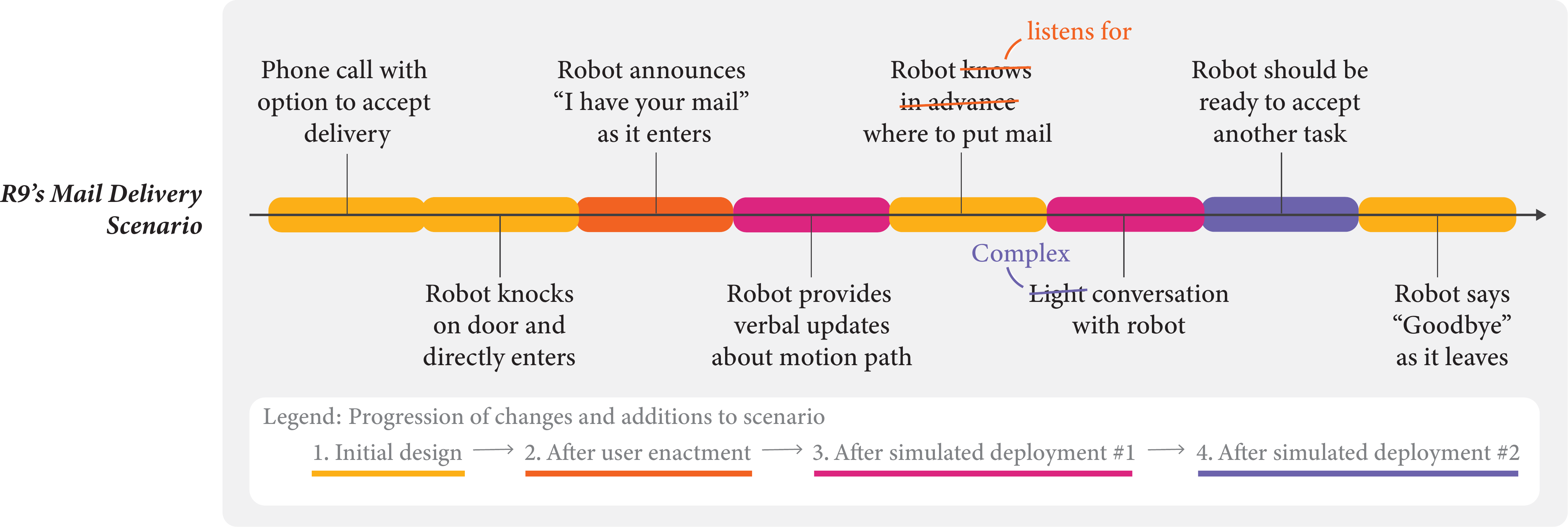}
    \caption{The evolution of an example design through Phase 1 and Phase 2 of \method{}. R9 designed a mail delivery scenario. Their initial design was simple, with limited robot interaction. However, repeated interactions with the robot allowed R9 to reflect on their preferences and iterate through different designs, adding steps or changing steps to make the robot behavior more appropriate. The final design includes verbal updates from the robot about its progress and deeper conversation with the robot.}
    \Description{Timeline showing the evolution of the mail delivery scenario designed by R9. The timeline shows segments of 4 colors reflecting the state of the scenario initially, after the user enactment, after the simulated deployment 1, and after the simulated deployment 2.
    Initially, the scenario starts with a phone call to announce the delivery, following by the robot knocking on door and directly entering. Then the robot should know in advance where to put the mail, before leaving. 
    After the user enactment, the robot needs to announce ``Ì have your mail'' as it enters and listen for the place where it should put the mail.
    After the first simulated deployment, the robot also need to provide verbal updates as it enters and conduct a light conversation before leaving.
    Finally, after the second simulated deployment, the robot should conduct a complex conversation instead of a light one and be ready to accept new tasks before leaving.}
    \label{fig:design_evolution}
\end{figure*}

\subsubsection{Insights from Engaging with Residents in Phase 2}
\label{phase2}

\paragraph{I4: Iterative interactions enable reflection on preferences for robot behavior.}
These repeated interactions with the robot throughout Phase 1 and Phase 2 allowed participants to reflect on their designs and make changes to how the robot should behave. Some participants, such as R2 and R3, made relatively few changes to their designs after the initial interaction. However, the remaining participants made significant changes as they realized their anticipated interaction with the robot was not what they actually desired.

R9's scenario evolution is visualized in Figure \ref{fig:design_evolution}. Initially, R9 was confident about how the robot should deliver the mail: no speech was necessary, and the robot should not do anything besides the mail delivery. However, after the first simulated deployment, R9 realized that due to ``the slowness of it'' and their apartment layout, they ``couldn't see'' what the robot was doing as it entered. In the first follow-up interview, R9 wanted to ``try'' getting verbal updates from the robot. R9 expressed, ``I don't know if I'll understand it,'' but they wanted to update the scenario design to try it. Then, at the end of the second simulated deployment, R9 further deviated from the original scenario by instructing the robot to do another task (\textit{i.e.}, deliver a note to the researchers). In the second follow-up interview, R9 commented, ``Having more visits made it smoother, easier.'' With the speech updates, R9 thought that the scenario ``worked out much better,'' but also that the robot should be ``made more personal by having conversation.'' Through repeat interactions with the robot, R9 reflected on and iterated through different designs to see what fit their preferences and needs. Generally, participants' initial impressions of what they wanted from the robot did not always match their true desires, which points to the importance of early, iterative experience with the robot under realistic conditions. 

\paragraph{I5: Experiencing the realistic scenario helps participants realize how it fits into their lives.}
Through repeated interaction, participants had time to reflect on the actual task the robot was doing. Since these interactions were as high-fidelity as possible, it provided context for participants to consider how that scenario fit into their life.

R8 initially asked the robot to deliver the morning newspaper. However, the newspaper arrived late, so the robot was also late with the delivery. 
After this experience, R8 voiced that the newspaper delivery was ``not a good task to set up for the robot.'' Instead, they wanted the robot to ``bring me ice for my afternoon cocktail.'' While ice delivery is a scenario that was not discussed during the interview, R8 independently imagined it after having the opportunity to reflect. For the second simulated deployment, the robot delivered the cup of ice, and R8 described the experience as ``wonderful.'' 
In the final reflective interview, R8 remarked that the process helped them ``conceptualize how it [the robot] could be a [...] very useful [...] tool for [...] people that are semi-confined.'' This example shows how experience with the robot and scenario under realistic conditions is a critical component to understand better what people want a robot to do and to conceptualize how it can be useful.

\paragraph{I6: Unexpected situations can appear from experiencing the robot in day-to-day life.}
Since the simulated deployments in Phase 2 were initiated by the robot without prior notice from the researchers, the interaction start was closer to what might happen in a real life experience. Participants were not specially prepared in the same way they made preparations to host the researchers for the session in Phase 1. Instead, we saw a snapshot of what might really happen when the robot completes the scenario in a deployment.

We experienced three unexpected situations from the simulated deployment. 
First, R5 had visitor when the robot arrived. The presence of the visitor changed the way the participant interacted with the robot, which subsequently changed how the robot needed to respond.
Second, R1 was using a scooter to move around, which we had not seen from previous visits. The scooter was parked in the way of where the robot needed to go, forcing the strategy to complete the handover to change.
Third, R8 was not wearing and consequently could not hear the robot speak at all.
These unplanned incidents show the need for flexible prototyping tools and for flexible systems in deployment because ``every day is different'' (R1). 

\paragraph{I7: Interacting with the robot without the mediation of the researchers can facilitate problem solving and idea generation.}
While participants were interacting with the robot, three ideas emerged that we had not considered during Phase 1 because the researchers were no longer present to mediate the interactions. 

First, R2 and R6 both felt that they helped the robot with its task. R2 recounts that they felt the robot ``didn't know how to get from there [the doorway] to here [the chair],'' so R2 ``helped'' the robot by ``putting [their] hand out [...] and he [the robot] came over.''
Similarly, R6 intervened as the robot attempted to place a tissue in the trash can because they ``realized there was a pole in the way'' that they thought the robot would ``run into.'' When asked about it, both participants were happy to help, and R6 was especially happy that the robot acknowledged their ``teamwork.''

Second, R7 expressed that they wanted validation and guidance from the robot. The robot was ``marvelous,'' but R7 felt ``inadequate'' to interact with it. Therefore, they wanted ``instructions'' from the robot during the interaction so that they knew what to do.

Third, R7 additionally mentioned that having the robot could help them to find a new role at the facility. They were interested to see if they could learn to use the robot, then help other residents learn as well. This new role could add value to their current life.

The lack of researchers forced residents to directly interact with the robot, causing them to consider the interaction with the robot instead of relying on the researchers' input as some did in Phase 1.


\subsubsection{Insights from Interacting with Caregivers in Phase 3}

\paragraph{I8: Familiarity with the robot helps shape caregiver expectations for what the robot can do.}
We tried to interview caregivers who had the opportunity to see the robot in action at the facility, which allowed for more concrete opinions on how this robot is perceived by the caregivers. While not all caregivers had seen the robot, all of them were familiar with it through hearing about it from either other facility staff or the residents. For example, C1 specifically commented that ``the size was not overly cumbersome.'' She further explained that seeing the robot in the facility was ``invigorating'' and made it ``not as leery or scary'' compared to when the abstract idea of a robot was first introduced and there were ``too many open questions.'' This excitement is encouraging as it supports the opportunity to use the robot in a future deployment phase.

\paragraph{I9: Common ground creates an environment where we can get meaningful feedback about the robot.}
Sharing our experiences from working with the residents and reflecting with the caregivers about their daily responsibilities built common ground that led to a mutual understanding of the challenges we were addressing. Common ground created the opportunity for caregivers to provide more relevant feedback. For example, C3 remarked that the tasks the residents designed ``would actually be perfect'' because she felt they would fit well with her needs as a caregiver. C3 further discussed the need for robots to be cognizant of resident preferences, emphasizing that ``every resident has their own preferences about how they like things.'' This feedback both confirms that the scenarios designed by participants are reasonable and also allows us to better understand how the robot can fit with the caregivers' needs and expectations.

\paragraph{I10: Discussion of the robot's role in assisted living elicits reflection on authority over the robot.}
At the end of the interview, we discussed more broadly about how the robot could fit into the assisted living environment. This discussion introduced a meaningful reflection about who should supervise and control the robot. All caregivers wanted some level of supervision but also felt that residents should be able to make requests from the robot. This issue of shared control led to C2 explaining the ``ethical question'' of how to ``preserve people's dignity and their ability to make choices'' while balancing what would be ``safest'' to do. Residents may have desires that do not align with their care needs, but it is not clear even among general caregiving practices how to balance care needs with resident wishes. 
\section{Discussion}
We proposed \method{} as a way to engage older adults in the design of assistive technologies and implemented \method{} in a case study with a robot in a senior living facility. Below, we revisit our research questions from \S\ref{rq} and follow with a general discussion of \method{}.

\subsection{Discussion of Research Questions}
\subsubsection{RQ1: How can designers effectively engage older adults in the design of assistive technologies?}
Our findings show that facilitating multiple high-fidelity interactions with the robot is an effective way to engage older adults. We observed that the emphasis on \textit{in situ} exploration of robot capabilities and enacting interactions with the real robot fostered engagement in Phase 1. Many participants were curious about the robot or eager to see if it would be able to help them with specific tasks. Prompting them step by step to provide general ideas and personal preferences about what the robot should do throughout the scenario helped them think through the interaction steps. Even if participants were unable to conjure abstractly what the robot should do, the enactment process facilitated idea generation by providing a natural prompt for them to react to---the robot's actions themselves. For example, when the robot extended its arm toward the participant to hand an item over, that participant was prompted to either extend their hands to accept the item, turn it away, or redirect the robot to perform a different action.

In Phase 2, we added further elements of realism by incorporating realistic task initiation, removing the researcher presence from the interaction, and asking the residents to simply use the robot (instead of acting). Whenever possible, the robot performed genuine, relevant tasks for them, such as delivering a real cup of ice that the participant then immediately used with their drink. Solving a real need that the resident had at that moment facilitated engagement, and it also helped to generate more concrete design recommendations from the residents. Facilitating multiple high-fidelity interactions for the older adults allowed them to better envision how the robot should fit into their daily lives and prompted them to reflect more critically on their experience with the robot.

\subsubsection{RQ2: How can designers better understand the challenges of integrating assistive technologies in genuine caregiving environments for older adults?}
Our findings show that demonstrating interaction designs \textit{in situ}, repeatedly experiencing these interactions, and integrating caregiver perspectives can all help build a better understanding of the challenges associated with integrating assistive technologies into care environments. First, the emphasis on demonstrating \textit{in situ} interactions with the real robot provided a new understanding of technical challenges and environmental considerations. For example, factors such as loud ambient noises (\textit{e.g.}, televisions or music) and the inability of some residents to speak loudly or clearly caused the built-in microphone on the robot to be unable to reliably capture the residents' speech. Such technical challenges would need to be addressed before the robot could be reliably deployed in a senior living facility.

Additionally, engaging residents over the course of multiple sessions on different days provided exposure to some unexpected situations that can arise in day-to-day life. For example, because the study sessions in Phase 2 were initiated without external warning from the researchers and the times were not always set in advance, we experienced situations that could have led to a breakdown based on the basic scenario design. For example, the robot once encountered another resident who was visiting our participant while it tried to deliver the mail, meaning the robot's behaviors and capabilities would need to accommodate an impromptu multi-party interaction. Although anticipating all possible situations is not feasible, our realistic interactions offered a glimpse of the types of emergent challenges the robot might face in a deployment.

Finally, feedback from the caregivers provided different perspectives on the challenges of integrating assistive technologies. While the caregivers agreed that the residents should be able to make requests from the robot, they felt that they needed high-level authority over the robot to ensure residents were not asking the robot for things that could cause them harm (\textit{e.g.}, an individual taking medication asking the robot for foods that might cause a drug interaction). Maintaining residents' autonomy versus supervising their choices is an open question even within conventional caregiving practices. 
Talking to the caregivers and residents separately provided valuable information, but considering how their perspectives fit together allows a more comprehensive view of integrating technology in daily activities and caregiving practices of older adults.

\subsection{Discussion of \method{}}
Overall, \method{} facilitated engagement with older adults and elicitation of considerations for integrating a robot into their daily lives. In the following paragraphs, we discuss the use of this method, focusing on its benefits, other scenarios where it may be applied, and how it fits into the wider context of system design and development.

\subsubsection{Benefits of \method{}}
We distilled our findings and contextualized them in the challenges discussed in \S\ref{intro}, resulting in five benefits that show the potential \method{} has for research with older adults:
\begin{enumerate}
    \item \textit{Promotion of inclusive and accessible design} ---
    Since \method{} is based on having participants interact with the target technology, limitations in participants' abilities to complete the study activities can help emphasize the necessary requirements for technology and interaction design. In addition, conducting the sessions in participants' living spaces allows individuals who are unable to travel to also participate. For example, four participants in our case study might not have been able to come to another study site or take part in some activities since two of them were manual wheelchair users and another two had dexterity impairments).
    \vspace{0.1cm}
    \item \textit{Better understanding of technology-environment fit by participants and researchers} ---
    The opportunity to explore the robot in the design phase and to experience the interaction during the simulated deployments provides \textit{participants} with a concrete idea of the robot's capabilities, which helps them ideate and refine what a robot can do for them. At the same time, \textit{researchers} can gain a better understanding of residents' lives, particularly how residents desire to interact with the system. For example, even though we observed some participants struggling to formulate how they desired the robot to interact, through \method{}, they were able to design acceptable scenarios (see insights I2, I3, I4, I5).
    \vspace{0.1cm}
    \item \textit{Vetting of designs under realistic conditions} ---
    As members of the target user population who have experienced the robot in genuine relevant use cases, residents' satisfaction with the system in the simulated deployment can serve as a predictor of the acceptance of the technology when it is deployed. For example, all of the residents except R4 were willing to interact with the robot, and most of them asked about the robot after the study concluded.
    \vspace{0.1cm}
    \item \textit{Early exposure to practical challenges and considerations} ---
    The simulated deployments allow researchers to assess the capabilities that the robot will need and test how well a current system is able to fulfill these requirements (\textit{e.g.}, navigation, grasping, social capabilities). Repeated interactions facilitate observation of uncommon situations, which may increase the robustness of the deployed systems. For example, as shown by insights I6 and I7, we were able to witness uncommon situations and assess what additional sensors and changes to modalities were required to interact efficiently.
    \vspace{0.1cm}
    \item \textit{Concrete, relevant feedback from other stakeholders} ---
    Engaging caregivers facilitated the assessment of the design ideas generated by older adults and the discovery of new design ideas, and it also raised considerations that residents may not have discussed. Due to the exposure to the robot and common ground developed through mutual sharing of experiences, caregivers could easily relate to our research, evaluate design ideas, and discuss the need for robot supervision (see insights I8, I9, and I10).
\end{enumerate}
We believe these benefits highlight the promise \method{} holds for designing with older adults. This method can offer benefits to other domains and populations as well, which we discuss below.

\subsubsection{Application to Other Domains and Technologies}
We believe that \method{} is not limited to robotics or older adults but has the potential to benefit the design of technology for other marginalized or vulnerable populations, \textit{e.g.}, children, individuals with cognitive impairment, individuals with blindness or visual impairment, or users with long-term physical disability. For example, certain activities such as cooking, navigation, home exercise, and tutoring are highly dependent on the specific ecosystem of use (\textit{e.g.}, home, community center, school). Designing assistive technologies, such as a smart cane, a cooking assistant, or a robotic walker, to help with such activities can benefit from \method. Introducing the technology early in its context of use and using simulated deployments can provide early and realistic feedback on the feasibility, accessibility, acceptability, and usability of the proposed ideas. 
We expect each phase of \method{} to need adaptations to fit the specific population, environment, technology, and use case being considered. For example, the other stakeholders in Phase 3 would change to family members in a home situation and to teachers and other students in a school setting. In settings that do not clearly involve other stakeholders, domain experts familiar with the vulnerable population (\textit{e.g.}, occupational therapists for blind individuals) can ensure that the designs would not interfere with other interventions or cause unintentional harm. 
Adapting \method{} to other emerging technologies and domains has the potential to provide similar benefits to what we experienced to design scenarios with an assistive robot, although future work is needed to understand the extent that these benefits translate.


\subsubsection{Considering the Bigger Picture}
\method{} fits within the wider context of assistive technology development as a design step to build toward a more autonomous system. While we used one cycle of \method{}, more cycles could be added to further improve and explore other aspects of the design. Each cycle can gradually increase the autonomy of the technology, building up to a fully functioning system. For example, we used a full WoZ setup, but next we could use a higher-level Wizard of Oz (WoZ) similar to work by \citet{senft2019teaching}, where the operator provides waypoints for navigation but still handles speech and manipulation. We could alternatively progress to include automation by the end users similar to work by \citet{winkle2021leador}. The advantage of iteratively increasing autonomy with \method{} is the increased confidence that the final system will succeed in a more in-depth evaluation or deployment.

\subsection{Limitations \& Future Work}
Our work has a number of limitations that point to future work, regarding \method{} and our case study, which we discuss below.

\paragraph{Methodology. } \method{} shows potential to help future researchers design scenarios with older adults, but it has three key limitations. First, it involves more setup work compared to other PD approaches. Using the WoZ approach to create realistic interactions means that we need an interface that allows full robot control. Nevertheless, developing this interface provides a starting point for future gradual automation of the system. Through WoZ, we could see what technical issues need to be addressed in future systems before investing the time to automate them. While more time is required up front, we expect that in the long term, it will shorten overall design and development time and lead to a more robust system.
Second, while the steps of \method{} generalize to other scenarios, \method{} has limited scalability due to the amount of scenario-specific setup involved (\textit{e.g.}, WoZ controls for the target system), and the design findings themselves do not necessarily generalize to other care settings and scenarios. 
Finally, the use of WoZ also introduces artifacts such as delays while the operator types speech for the robot, which might limit the quality of the participants' feedback. 

\paragraph{Case study. } Our case study using \method{} has four key limitations that should be addressed with future work. 
First, we only engaged with a subset of residents and caregivers. Because participants had to volunteer, it is possible that they represent a more optimistic and accepting view than other individuals who declined participation. Further, we only worked with residents who had the capacity to provide informed consent. Therefore, we did not work with participants with severe cognitive impairments, which excludes many individuals in assisted living. Future work should seek to engage a wider pool of participants to investigate how \method{} can be applied to address other challenges.
Second, our case study included a relatively small number of participants with only one cycle of \method{}. While this configuration already demonstrated the potential of the method, future work should investigate more long-term effects. The novelty of the robot may wear off over time, and the patterns and preferences of residents may also keep changing over longer exposure to the system. 
Third, our study involved a single robot platform, which was selected as it provides the required capabilities at a low price point and is designed to work in home settings. Although \method{} is designed to evaluate a single platform, \method{} could be used with other platforms or be combined with other PD work \cite[e.g.,][]{bradwell2021user} to explore trade offs and preferences for different platforms and capabilities.
Finally, our participants only included residents and caregivers. Future work should incorporate other stakeholders, such as family and other facility staff, into the different design phases to increase the ecological validity of the resulting designs.
\section{Conclusion}
This paper presented \methodfull{} (\method{}), a participatory design method crafted to fully engage older adults in the design of assistive technologies and to understand challenges with deploying them in senior living facilities. \method{} involves three phases: (1) a \textit{co-design} phase to explore the technology and design an initial scenario; (2) a \textit{simulated deployment} phase to evaluate the scenario in realistic conditions; and (3) an \textit{interview} phase to reflect on the resulting scenarios with other stakeholders (\textit{e.g.}, caregivers). We applied \method{} to a case study that involved co-designing interactions between older adults and an assistive robot with residents and caregivers of a senior living facility. We found that the residents have a wide range of needs and preferences that affect how robots and interactions should be designed, such as privacy considerations when gaining entry to the resident's space or social considerations about how much the robot should talk throughout the scenario. 
Our case study revealed a number of insights into \method{}, which help us understand its benefits and limitations. Overall, \method{} creates an immersive, realistic, and reflective co-design experience that benefits both participants and researchers. With a strong focus on the situating interactions within the realistic ecosystem, \method{} facilitates engagement with older adults by considering their cognitive and physical abilities. Researchers have the additional benefit of early exposure to technical and behavioral challenges that need to be addressed prior to deployment. 
While \method{} was developed to address the challenges of working with older adults, we believe that it has application to other domains and technologies and that it could help designers and researchers more deeply engage vulnerable and marginalized communities in the design of assistive technologies.

\begin{acks}
We would like to thank the residents and caregivers for participating in our study. 
This material is based upon work supported by a University of Wisconsin--Madison Vilas Associates Award, National Science Foundation (NSF) award IIS-1925043, and the NSF Graduate Research Fellowship Program under Grant No. DGE-1747503. 
Any opinions, findings, and conclusions or recommendations expressed in this material are those of the authors and do not necessarily reflect the views of the NSF.
\end{acks}
\balance
\bibliographystyle{ACM-Reference-Format}
\bibliography{biblio}


\end{document}